\documentclass[11pt]{article}

\usepackage[]{emnlp2021}

\usepackage{times}
\usepackage{latexsym}

\usepackage[T1]{fontenc}

\usepackage[utf8]{inputenc}

\usepackage{microtype}
\usepackage{comment}

\usepackage{amsfonts}
\usepackage{algorithm}
\usepackage{amsmath}
\usepackage[noend]{algpseudocode}
\usepackage{hyperref}
\usepackage{booktabs}
\usepackage{graphicx}
\usepackage{enumitem}
\usepackage{arydshln}
\usepackage{pdfpages}
\usepackage{pgfplots}

\title{Topic-Aware Contrastive Learning for Abstractive Dialogue Summarization}

\author{
    Junpeng Liu \textsuperscript{\rm 1}{\thanks{\ \ Work done at JD.com. }},
    Yanyan Zou\textsuperscript{\rm 2},
    Hainan Zhang\textsuperscript{\rm 2},
    {\bf Hongshen Chen\textsuperscript{\rm 2}}, \\
    {\bf Zhuoye Ding\textsuperscript{\rm 2}},
    {\bf Caixia Yuan\textsuperscript{\rm 1}}
    {\bf \and Xiaojie Wang\textsuperscript{\rm 1}} \\
    
    \textsuperscript{\rm 1}Beijing University of Posts and Telecommunications, Beijing, China\\
    \textsuperscript{\rm 2}JD.com, Beijing, China \\
    \texttt{\{jeepliu, yuancx, xjwang\}@bupt.edu.cn} \\
    \texttt{\{zouyanyan6,dingzhuoye\}@jd.com} \\ 
    \texttt{zhanghainan1990@163.com, ac@chenhongshen.com}
}

\begin{document}
	\maketitle
	\begin{abstract}
		Unlike well-structured text, such as news reports and encyclopedia articles, dialogue content often comes from two or more interlocutors, exchanging information with each other.
		In such a scenario,  the topic of a conversation can vary upon progression and the key information for a certain topic is often scattered across multiple utterances of different speakers, which poses challenges to abstractly summarize dialogues.
		To capture the various topic information of a conversation and outline salient facts for the captured topics, this work proposes two topic-aware contrastive learning objectives, namely coherence detection and sub-summary generation objectives, which are expected to implicitly model the topic change and handle information scattering challenges for the dialogue summarization task.
		The proposed contrastive objectives are framed as auxiliary tasks for the primary dialogue summarization task, united via an alternative parameter updating strategy.
		Extensive experiments on benchmark datasets demonstrate that the proposed simple method significantly outperforms strong baselines and achieves new state-of-the-art performance.
		The code and trained models are publicly available via \href{https://github.com/Junpliu/ConDigSum}{https://github.com/Junpliu/ConDigSum}.

	\end{abstract}
	
	\section{Introduction}
	\label{sec:intro}
	Online conversations have become an indispensable manner of communication in our daily work and life.
	In the era of information explosion, it is paramount to present the most salient facts of conversation content, rather than lengthy utterances, which is useful for online customer service \cite{liu2019automatic} and meeting summary \cite{zhao2019abstractive}. 
	This work focuses on \textit{abstractive dialogue summarization}.
	To summarize dialogues, one simple way is to directly apply existing document summarization models to dialogues \cite{shang2018unsupervised,gliwa19} or to employ hierarchical models to capture features from different turns of different speakers \cite{zhao2019abstractive,zhu2020hierarchical}.
	However, succinctly summarizing the dialogue is much more challenging.

	\begin{figure}[t]
        \centering
        \setlength{\abovecaptionskip}{0.2cm}
        \includegraphics[width=0.5\textwidth]{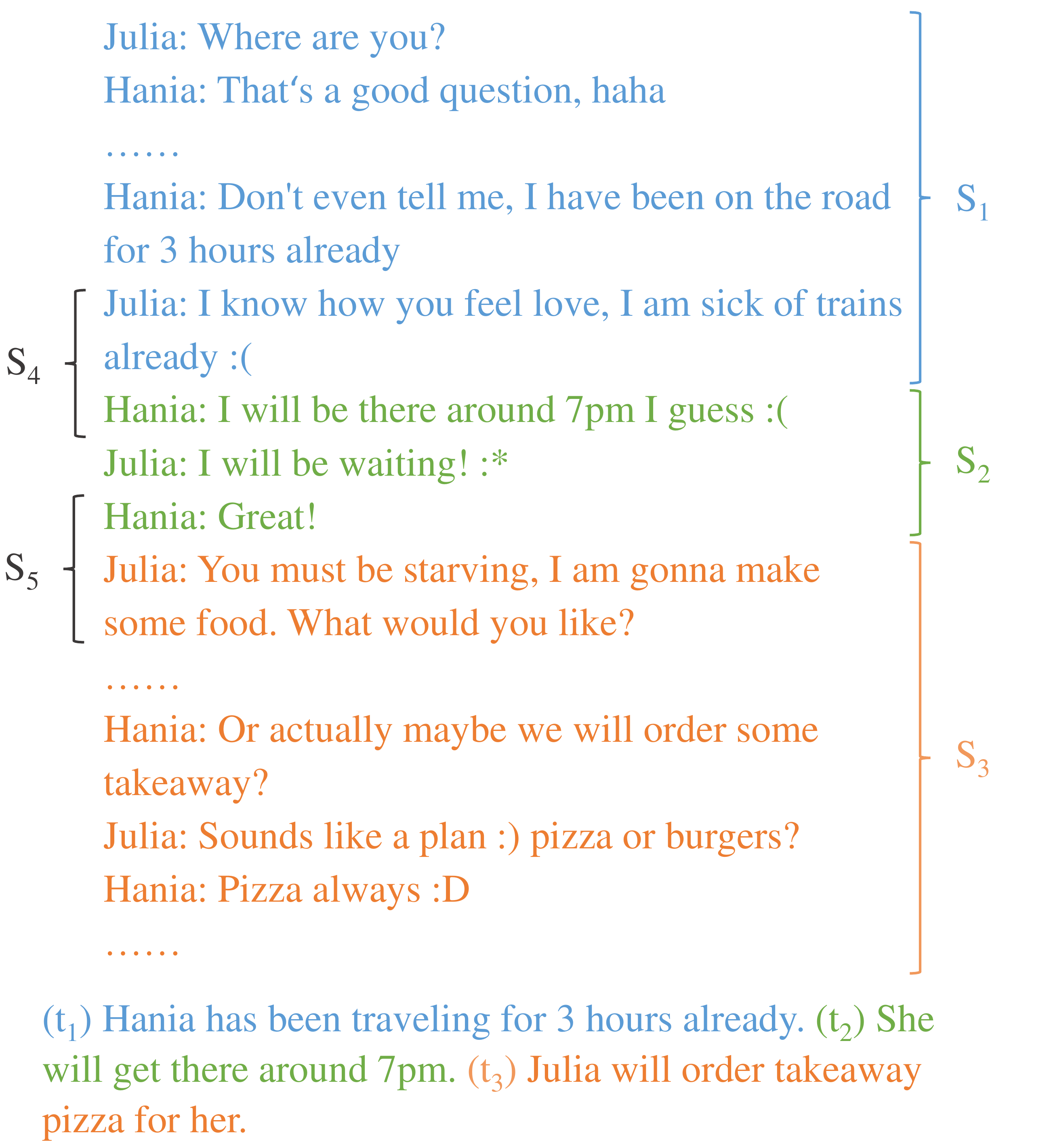}
        \caption{A dialogue and its paired summary. $S_1$, $S_2$, and $S_3$ stands for referred topic snippets, \textit{current situation}, \textit{time of arrival} and \textit{food to eat}, respectively. The corresponding summary consists of three sentences $t_1$, $t_2$ and $t_3$. Each $t_i$ corresponds to one snippet $S_i$ ($i=1, 2, 3$). $S_4$ and $S_5$ are inter-topic snippets. }
        \vspace{-2mm}
        \label{Fig:intro-example}
        \vspace{-4mm}
    \end{figure}

	

    The well-structured textual descriptions, such as news reports \cite{see2017get} and academic papers \cite{nikolov2018datadriven}, often come from one single speaker or writer where the information flow is more natural and clearer with paragraphs and sections.
	Differently, consisting of multiple utterances from two or more interlocutors, the conversational content is in a complicated flow with information exchange and the focused topic can vary upon the conversation progression.
	On the other hand, the salient information for a specific topic is often scattered across multiple utterances and can be presented separately.  
	Exemplified by Figure \ref{Fig:intro-example}, this dialogue touches three topics, \textit{current situation, time of arrival} and \textit{food to eat}, where the corresponding topic snippets are $S_1$, $S_2$ and $S_3$, respectively.
	The central ideas of each topic is summarized with one sentence, covering information from multiple utterances, i.e., $t_1$ for $S_1$, $t_2$ for $S_2$, and $t_3$ for $S_3$.
	We also observe that utterances residing in the same topic (e.g., $S_1$, $S_2$ and $S_3$) is inherently more coherent than those coming from different topics (e.g., the inter-topic snippet $S_4$ and $S_5$), which reveals the underlying relationships between topic and utterance coherence, also demonstrated by \citet{glavavs2020two}.

	Recent studies involves intrinsic information of dialogues to handle the challenges for summarizing dialogues, such as topic segment features \cite{Liu2019TopicAwarePN,li2019keep,chen2020multi}, dialogue acts \cite{goo2018abstractive} and conversation stages \cite{chen2020multi}.
	Although such existing models have demonstrated the effectiveness of the dialogue analysis on generating summaries, additional human efforts in data annotations or extra topic segmentation algorithms are necessary.
	For example, \citet{goo2018abstractive,liu2019automatic} require extensive expert annotations on dialogue acts, while the knowledge of visual focus of each speaker and topic segment is a must for \citealp{li2019keep}'s work, which are both expensive and sometimes hard to obtain. 
	\citet{Liu2019TopicAwarePN,chen2020multi} need extra algorithms to obtain topic segment information, which works with the primary summarization model in a pipeline manner and thus may cause error propagation.
	Different from the structured text where a paragraph or a section can be treated as natural topic segment, it is difficult to accurately segment topics of dialogues.

    Recall the inherent relationships between the topic and utterance coherence, this work proposes to implicitly capture the dialogue topic information by modeling the utterance coherence in a contrastive way.
    The coherence detection objective is constructed to push the model to focus more on snippets that are more coherent and likely contain salient information from the same topics. 
    Further, since we aim to generate better summaries for each topic in a dialogue, we also introduce the \textit{sub-summary generation objective}, which is expected to force the model to identify the most salient information and generate corresponding summaries.
    Note that both objectives are constructed in a contrastive way where no additional human annotations or extra algorithms are required.
    Such two contrastive objectives can be coupled with the primary dialogue summarization task via an alternating parameter updating strategy, resulting in our final model \textsc{ConDigSum}.
    Experiments on two dialogue summarization datasets demonstrate the effectiveness of our proposed contrastive learning objectives for dialogue summarization which achieves new state-of-the-art performances.




	\section{Proposed Method}
	
	\begin{figure}[tbp] 
        \centering 
        \includegraphics[width=0.5\textwidth]{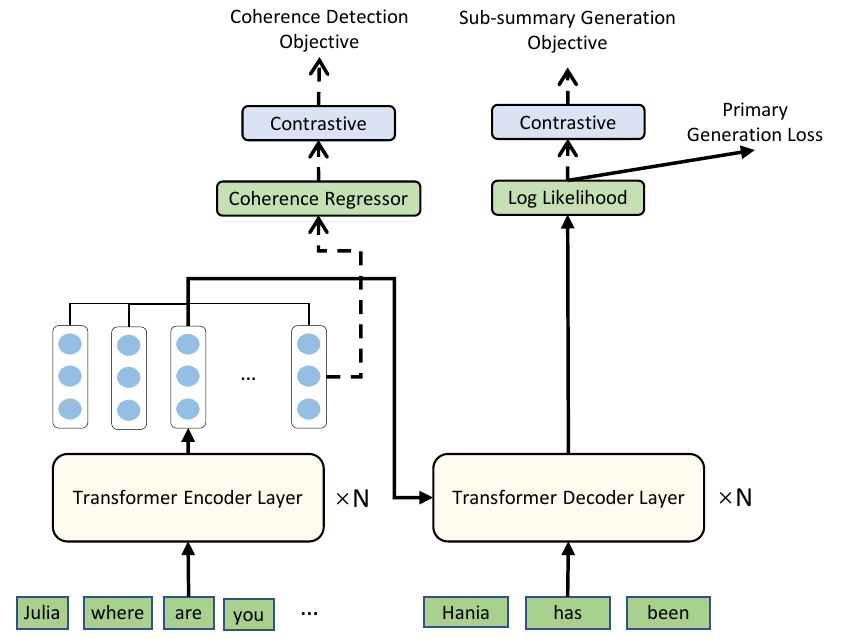} 
        \caption{Model structure with contrastive objectives. } 
        \label{Fig:model} 
        \vspace{-5mm}
    \end{figure}
	
	\subsection{Sequence-to-Sequence Learning}
    \label{sec:seq2seq}
	In this work, we frame the abstractive summarization task as a sequence-to-sequence learning problem.
	The sequence-to-sequence Transformer \cite{vaswani2017attention} is adopted as our backbone architecture, where the model takes as input the dialogue utterances and generates a corresponding summary.
	Specifically, given a dialogue $ \mathcal{D} = (u_1, u_2,..., u_{|\mathcal{D}|})$, consisting of $ |\mathcal{D}| $ utterances, coupled with its corresponding summary $ T_{\mathcal{D}} = (y_1, y_2,..., y_{|T_{\mathcal{D}}|}) $ in the length of $|T_{\mathcal{D}}|$, the goal is to learn the optimal model parameters $\theta$ and to minimize the negative log-likelihood:
	\begin{gather}
		\label{equ:seq2seq}
\mathcal{L}^{\mathcal{D},T_{\mathcal{D}}} = \sum_{i=1}^{|T_{\mathcal{D}}|} -\log p(y_i | y_{1:i-1}, \mathcal{D};\theta )
	\end{gather}
	where $y_{1:i-1}$ denotes the first $i-1$ tokens of the output sequence (i.e., $ y_{1:i-1} = (y_1, y_2,...,y_{i-1}) $).
	For a certain batch of dialogue-summary pairs $ \mathcal{B} = (\langle \mathcal{D}_1, T_{\mathcal{D}_1}\rangle,\langle \mathcal{D}_2, T_{\mathcal{D}_2}\rangle, \dots, \langle \mathcal{D}_{|\mathcal{B}|}, T_{\mathcal{D}_{|\mathcal{B}|}}\rangle)$, the negative log-likelihood is calculated as:
	\begin{equation}
		\label{equ:seq2seq_loss}
		\mathcal{L}_{main}^{\mathcal{B}} = \frac{1}{|\mathcal{B}|} \sum_{\langle \mathcal{D}, T_{\mathcal{D}}\rangle \in \mathcal{B}} \mathcal{L}^{\mathcal{D},T_{\mathcal{D}}} 
	\end{equation}

	\subsection{Contrastive Objectives}
	\label{sec:task}
	In this section, we introduce two contrastive objectives, coherence detection and sub-summary examination objectives, which can be considered as auxiliary tasks during training phase and reinforce the primary dialogue summarization task.

	\paragraph{Coherence Detection Objective.}
	\label{sec:aux_cor_task}
    The access to topic labels of dialogues often requires extra expert annotations or additional topic segment algorithms, which is expensive or may introduce error propagation.
	Considering the observation that text coherence is inherently related to the text topic (refer to Section \ref{sec:intro}), instead, we obtain the topical information of a dialogue by modeling the coherence change among utterances.
	The assumption behind this is that utterances within the same topic are more coherent than those spanning across different topics, based on which we construct the contrastive \textit{coherence detection objective}. 
	
	To conduct contrastive learning, we construct positive-negative pairs with self-supervision.
	Recall that a dialogue consists of $|\mathcal{D}|$ utterances, i.e.,  $\mathcal{D} = ( u_1, u_2, \dots, u_{|\mathcal{D}|} )$.
	We introduce a window comprising a subsequence of $k$ $(k<|\mathcal{D}|)$ utterances of a dialogue $\mathcal{D}$, as a \textit{snippet}, denoted as $\mathcal{S}_k^{\mathcal{D}}$.  
	For instance, $(u_j, u_{j+1}, \dots, u_{j+k})$ is an example snippet for dialogue $\mathcal{D}$ where $j \in [1, |\mathcal{D}|-k]$ is an integer utterance index.
    Such a snippet is regarded as a positive example, while the corresponding negative snippet $\widetilde{\mathcal{S}_k^{\mathcal{D}}}$ is constructed by shuffling the order of  sentences inside $\mathcal{S}_k^{\mathcal{D}}$.
    Given a pair of positive and negative examples, denoted as $\mathcal{P}_{co}^{\mathcal{D}}=(\mathcal{S}_k^{\mathcal{D}}, \widetilde{\mathcal{S}_k^{\mathcal{D}}})$, the contextual representations of each snippet can be obtained through the last layer of the Transformer encoder, denoted as $E_{\mathcal{S}_k^{\mathcal{D}}} $, $ E_{\widetilde{\mathcal{S}_k^{\mathcal{D}}}}$, individually.
    Then we can calculate the coherence scores within a snippet by:
    \begin{equation*}
    	y_{\mathcal{S}_k^{\mathcal{D}}} = {w_1} * E_{\mathcal{S}_k^{\mathcal{D}}}+ b_1; \quad y_{\widetilde{\mathcal{S}_k^{\mathcal{D}}}} = {w_1} * E_{\widetilde{\mathcal{S}_k^{\mathcal{D}}}} + b_1
    \end{equation*}
    where $ w_1 \in \mathbb{R}^d $ and $ b_1 \in \mathbb{R} $ are trainable parameters besides the original Transformer architecture, as depicted as \textit{Coherence Regressor} in Figure \ref{Fig:model}.
	The normalization with a softmax layer is conducted to obtain the final coherence score:
	\begin{equation*}
		[co(\mathcal{S}_k^{\mathcal{D}}), co(\widetilde{\mathcal{S}_k^{\mathcal{D}}})] = softmax([y_{\mathcal{S}_k^{\mathcal{D}}}, y_{\widetilde{\mathcal{S}_k^{\mathcal{D}}}}])
	\end{equation*}
	
	For a dialogue $\mathcal{D}$, there exist at least $|\mathcal{D}-k|$ contrastive snippet pairs, while, for simplicity, we randomly select $N_{co}<|\mathcal{D}-k|$ pairs for each epoch during training.
	The contrastive margin-based coherence loss is then calculated as:
	\begin{gather*}
	\mathcal{L}_{co}^{\mathcal{D}} = \\
	\frac{1}{N_{co}}\sum_{n=1}^{N_{co}}\max(0, {\delta}_{co} - ( co(S_{k,n}^{\mathcal{D}}) - co(\widetilde{S_{k,n}^{\mathcal{D}}}) ))
	\end{gather*}
	where $\delta_{co}$ is a margin coefficient by which we expect that the coherence score for the positive snippet is larger than the score for the negative one.
	$k$, $N_{co}$ and $\delta_{co}$ are hyperparameters.
	For a certain batch of dialogue-summary pairs $ \mathcal{B} = (\langle \mathcal{D}_1, T_{\mathcal{D}_1}\rangle,\langle \mathcal{D}_2, T_{\mathcal{D}_2}\rangle, \dots, \langle \mathcal{D}_{|\mathcal{B}|}, T_{\mathcal{D}_{|\mathcal{B}|}}\rangle)$, the margin-based contrastive loss is calculated as:
		\begin{equation}
		\mathcal{L}_{co}^{\mathcal{B}} = \frac{1}{|\mathcal{B}|} \sum_{\langle \mathcal{D}, T_{\mathcal{D}}\rangle \in \mathcal{B}} \mathcal{L}_{co}^{\mathcal{D}}
	\end{equation}
	In this setting, we only use the dialogue while the summary is untouched.
	The coherence loss can be used to update the parameters in the encoder.

	\begin{algorithm}[t]
	\renewcommand{\algorithmicrequire}{\textbf{Input:}}
	\renewcommand{\algorithmicensure}{\textbf{Output:}}				
	\caption{Snippet selection for a sub-summary}
	\label{alg:sub-summary}
	\begin{algorithmic}
		\Require A sub-summary $ t_i \in T $, a dialogue $\mathcal{D}$ containing $|\mathcal{D}|$ utterances, sliding window size interval $\left[a,b\right]$
		\Ensure  ($ S_\text{pos}^i ,S_\text{neg}^i$) for $ t_i $
		\State $\mathcal{W} = \emptyset$
		\For {$w = a$ to $b$}
		\For {$j = 1$ to $|\mathcal{D}|-w$}
		\State $\text{cand} = \mathcal{D}_{j, j+w}$
		\State $ r(j,w) \gets \texttt{ROUGE}(\text{cand}, t_i) $ 
		\State $\mathcal{W} \gets \mathcal{W} \cup \text{cand}$
		\State $j \gets j+w/2$
		\EndFor
		\State$ w \gets w+1$
		\EndFor
		\State $ j_\text{best}, w_\text{best} \gets \mathop{\arg\max}_{j, w} r(j,w)$
		\State $ S_\text{pos}^i \gets \mathcal{D}_{j_\text{best},(j_\text{best}+w_\text{best})} $
		\State $S_\text{neg}^i \gets \mathcal{W} \setminus S_\text{pos}^i$
	\end{algorithmic}
	\end{algorithm}
	
		\paragraph{Sub-summary Generation Objective.} The summary of a long dialogue always consists of multiple sentences each of which is regarded as a \textit{sub-summary}.
	Considering the fact that one dialogue may contain more than one topics, we assume that each sub-summary is related to one topic.
	Hence, we introduce the contrastive \textit{sub-summary generation objective}.
	
	It is straightforward to obtain the sub-summaries by dividing the whole summary into single sentences via period symbols\footnote{More details are in the appendix.}.
	For simple illustration, here we denote the corresponding target summary of a dialogue $\mathcal{D} = ( u_1, u_2, \dots, u_{|\mathcal{D}|} )$ as $T_{\mathcal{D}} = (t_1, t_2, ..., t_m)$, where $m$ is the number of sentences and each $t_i$ is considered as a sub-summary.
	Given a sub-summary $t_i$, we can retrieve the most related snippet $\mathcal{S}_{\text{pos}}^{i}$ from the dialogue $\mathcal{D}$ according to the ROUGE-2 recall score \cite{lin2004rouge}. 
	The detailed selection algorithm is presented in Algorithm \ref{alg:sub-summary}.
	Given an integer window size $w \in \left[a, b\right] (0< a \leq b < |\mathcal{D}|)$, we can slide the window over the dialogue $\mathcal{D}$ in the stride of half window size and obtain a set of candidate snippets $\mathcal{W}$.
	Enumerating each snippet candidate in $\mathcal{W}$ and calculating the ROUGE-2 recall score with the sub-summary $t_i$, we can get the optimal snippet scored the highest, which is selected as the most related snippet and regarded as the positive example $S_\text{pos}^i $.
	The corresponding negative example is randomly picked from the rest snippets in $\mathcal{W}$, denoted as $S_\text{neg}^i$.
	Now, we have constructed the contrastive sub-summary generation pairs $\{(S_\text{pos}^i, t_i), (S_\text{neg}^i, t_i)\}$.
	Like the primary dialogue summarization task, we also model the sub-summary generation objective as a sequence-to-sequence learning problem. 
	Following Equation~\ref{equ:seq2seq}, the negative log-likelihoods are calculated as:
	\begin{gather*}
		\mathcal{L}_{pos}^{t_i} = -\log(\prod_{j=1}^{|t_i|} p(t_j^i | t_{1:j-1}^i, \mathcal{S}_{\text{pos}}^i;  \theta)) \\
		\mathcal{L}_{neg}^{t_i} = -\log(\prod_{j=1}^{|t_i|} p(t_j^i | t_{1:j-1}^i, \mathcal{S}_{\text{neg}}^i; \theta))
	\end{gather*}
	where $t_j^i$ refers to the $j_{\text{th}}$ token in $t_i$ and $t_{1:j-1}^i$ stands for all preceding tokens before position $j$.
	The normalized scores after the softmax layer can be regarded as the irrelevance score to show how irrelevant a snippet is to a sub-summary:	
	\begin{gather*}
	[su(S_{\text{pos}}^i), su(S_{\text{neg}}^i)] = softmax([\mathcal{L}_{pos}^{t_i}, \mathcal{L}_{neg}^{t_i}]) 
    \end{gather*}
	For a dialogue $\mathcal{D}$ paired with its summary $T_{\mathcal{D}}$, at least $m$ contrastive pairs can be constructed, while, similar to the coherence case, we randomly select $N_{su} < m$ pairs for each epoch during training phase.
	Thus, we can construct a contrastive margin-based loss for dialogue $\mathcal{D}$:
	\begin{gather*}
	\mathcal{L}_{su}^{\mathcal{D},T_{\mathcal{D}}} = \\ \frac{1}{N_{su}}\sum_{n=1}^{N_{su}}\max(0, \delta_{su} - (su(S_{\text{neg}}^n) - su(S_{\text{pos}}^n)))
\end{gather*}
	where $\delta_{su}$ is a margin coefficient by which we would like the relevance score between a positive snippet and a sub-summary to be at least larger than the relevance score of the negative pair. 
    $a$, $b$, $N_{su}$ and $\delta_{su}$ are hyperparameters.
	For a certain batch of dialogue-summary pairs $ \mathcal{B} = (\langle \mathcal{D}_1, T_{\mathcal{D}_1}\rangle,\langle \mathcal{D}_2, T_{\mathcal{D}_2}\rangle, \dots, \langle \mathcal{D}_{|\mathcal{B}|}, T_{\mathcal{D}_{|\mathcal{B}|}}\rangle)$, the negative log-likelihood is calculated as:
	\begin{gather}
		\label{equ:subsummary}
		\mathcal{L}_{su}^{\mathcal{B}} = \frac{1}{|\mathcal{B}|} \sum_{\langle \mathcal{D}, T_{\mathcal{D}}\rangle \in \mathcal{B}} \mathcal{L}_{su}^{\mathcal{D},T_{\mathcal{D}}} 
	\end{gather}
    The sub-summary objective can be used to update the parameters in the encoder and decoder.

 	\begin{algorithm}[t]
	\renewcommand{\algorithmicrequire}{\textbf{Input:}}
	\renewcommand{\algorithmicensure}{\textbf{Output:}}
	\caption{Alternating Updating Strategy}
	\label{alg:alter}
	\begin{algorithmic}[1]
		\Require A batch of dialogue-summary instances $ \mathcal{B}$
		\Statex Coherence Task
		\State $ \mathcal{L}_{co}^{\mathcal{B}} = \frac{1}{|\mathcal{B}|} \sum_{\langle \mathcal{D}, T_{\mathcal{D}} \rangle   \in \mathcal{B}} \mathcal{L}_{co}^{\mathcal{D}} $
		\State $ \theta  \gets  \theta - \alpha w_{co} \frac{\partial L_{co}^\mathcal{B}}{\partial \theta}$
		\Statex Sub-summary Task
		\State $ \mathcal{L}_{su}^{\mathcal{B}} = \frac{1}{|\mathcal{B}|} \sum_{\langle \mathcal{D}, T_{\mathcal{D}} \rangle   \in \mathcal{B}} \mathcal{L}_{su}^{\mathcal{D}, T_{\mathcal{D}}} $
		\State $ \theta \gets  \theta - \alpha w_{su} \frac{\partial \mathcal{L}_{su}^\mathcal{B}}{\partial \theta}$
		\Statex Main Task
		\State $ \mathcal{L}_{main}^{\mathcal{B}} = - \frac{1}{|\mathcal{B}|} \sum_{\langle \mathcal{D}, T_{\mathcal{D}} \rangle  \in \mathcal{B}} \mathcal{L}^{\mathcal{D},T_{\mathcal{D}}}  $
		\State $ \theta  \gets  \theta - \alpha w_{main} \frac{\partial \mathcal{L}_{main}^\mathcal{B}}{\partial \theta}$
	\end{algorithmic}
\end{algorithm}

	\subsection{Multi-Task Learning}
	\label{sec:multi_task}
	The proposed two contrastive objectives can contribute to the primary dialogue summarization task during training phase, acting as auxiliary tasks.
	There are two options to combine the primary and auxiliary tasks: 1) summing the three objectives as a single one and update the model parameters using the summation loss; 2) alternatively update the model parameters using one of three objectives at each time.
	The empirical studies (Section \ref{subsec:eva}) show that the alternating updating strategy performs better. 
	Thus, in this work, we adopt the alternating parameter updating strategy, as shown in Algorithm \ref{alg:alter}. 
	For a certain batch of dialogue-summary pairs, three objectives are adopted to update parameters in sequence.
	We first update the model parameters using the coherence objective, followed by the sub-summary and the primary generation objectives.
	The three objectives share the same learning rate $\alpha$.
	Since the main focus is to generate better dialogue summaries with the help of auxiliary contrastive objectives, we give more attentions to the primary task.
	Inspired by \citealp{dasgupta2016leveraging}, to drive the auxiliary tasks to contribute to the primary one yet not to be dominate, we also introduce task-wise coefficients to each task, denoted as $w_{co}$, $w_{su}$ and $w_{main}$, individually.
	Following experiments demonstrate the effectiveness of the alternating strategy and the introduced task-wise coefficients.

	\section{Experiment}
	\label{sec:exp}

	\subsection{Datasets}
	
	\paragraph{SAMSum} contains natural message-like dialogues in English written by linguists, each of which is annotated with summary by language experts \cite{gliwa19}.
	There are 14,732 dialogue-summary pairs for training, 818 and 819 instances for validation and test, respectively.

	\paragraph{MediaSum} is a large-scale dataset for dialogue summarization, containing interview transcripts collected from National Public Radio (NPR) \footnote{\href{www.npr.org}{www.npr.org}} and CNN \footnote{\href{www.transcripts.cnn.com}{www.transcripts.cnn.com}}, where the overview descriptions or discussion guidelines, coming with the transcripts, are considered as corresponding abstractive summaries \cite{zhu2021mediasum}.
	The whole corpus contains 463.6K instances, with 10K each for validation and testing individually, and the rest is for training.

	\subsection{Implementation Details}
	As mentioned in Section \ref{sec:seq2seq}, the sequence-to-sequence Transformer model is adopted as our backbone architecture, implemented using Fairseq toolkit\footnote{We empirically observed that different frameworks (e.g. Fairseq and Huggingface Transformer) may obtain different results under the same hyperparameter settings.} \citep{ott2019fairseq}. 
	To be specific, our model is initialized with a pre-trained sequence-to-sequence, i.e., BART \cite{lewis2020bart}.
	Thus they share the same architectures, 6-layer encoder-decoder Transformer for $\text{BART}_\text{BASE}$ and 12-layer Transformer for $\text{BART}_\text{LARGE}$.
	Each layer in $\text{BART}_\text{BASE}$ has 16 attention heads, and the hidden size and feed-forward filter size is 1024 and 4096, respectively, resulting in 140M trainable parameters. 	
	Each layer in $\text{BART}_\text{LARGE}$ has 16 attention heads, and the hidden size and feed-forward filter size is 1024 and 4096, respectively, resulting in 400M trainable parameters. 
	The dropout rates for all layers are set to 0.1.
	The optimizer is Adam \cite{kingmaadam} with warmup.
	The learning rate $\alpha$ for SAMSum is 4e-5, 2e-5 for MediaSum.
	The maximum number of tokens for a certain batch is 800 and 1100 for SAMSum and MediaSum, individually.
   	The margin coefficients $\delta_{co}$ and $\delta_{su}$ for the two contrastive objectives are always set to 1.
  	Other hyper-parameters of our methods, including $ w_{co}, w_{su}, k, a, b $ are tuned on the validation set. 
  	
%
%
%
%
    More implementation details and sensitivity tests for hyper-parameters are included in the appendix. 
    
    \subsection{Evaluation}
    	\label{subsec:eva}
	To evaluate our models, we utilized the ROUGE \cite{lin2004rouge} to measure the quality of summary output generated by different models.
	We adopted the files2rouge\footnote{\href{https://github.com/pltrdy/files2rouge}{https://github.com/pltrdy/files2rouge} Note that the ROUGE scores might vary with different tookits.} package based on the official \texttt{ROUGE-1.5.5.pl} perl script to get full-length ROUGE-1, ROUGE-2 and ROUGE-L F-measure scores.
	The recent popular automatic evaluation metric for text generation, BERTScore \citep{zhang2020bertscore}, is also presented for comparisons\footnote{We use version 0.3.8, with default English setting (roberta-large\_L17\_no-idf\_version=0.3.8(hug\_trans=4.4.0)-rescaled).}. 
	For simiplicity, we use R-1, R-2, R-L and \textsc{BertS} to refer to ROUGE-1, ROUGE-2, ROUGE-L and {BERTScore}, respectively.

\begin{table}[t]
		\centering
		\scalebox{0.85}{
			\begin{tabular}{lcccc}
				\toprule
				{Model} & {R-1} & {R-2} & {R-L} & {\textsc{BertS}} \\
				\midrule
				$\ast$Lead3                               & 31.4 & 8.7 & 29.4  & - \\
				$\ast$PTGen                    & 40.1 & 15.3 & 36.6 & - \\ 
				$\ast$DynamicConv + GPT-2               & 41.8 & 16.4 & 37.6 & - \\ 
				$\ast$FastAbs-RL              & 42.0 & 18.1 & 39.2 & - \\ 
				$\ast$DynamicConv + News              & 45.4 & 20.7 & 41.5 & - \\ 
				Multiview BART               & 53.9 & 28.4 & 44.4 & 53.6 \\ 
				\midrule
				$\ast$$\text{BART}_\textsc{base}$   & 46.1 & 22.3 & 36.4 & 44.8 \\ 
				$\ast$$\text{BART}$  & 52.6 & 27.0 & 42.1 & 52.1 \\ 
				$\ast$$\text{BART}_\textsc{ori}$        & 52.6 & 27.2 & 42.7 & 52.3\\  
               $\textsc{ConDigSum}_\text{BASE}$ & 48.1 & 24.0 & 39.2 & 48.0\\
				\textsc{ConDigSum}                                 & \textbf{54.3 }& \textbf{29.3} & \textbf{45.2} & \textbf{54.0} \\
                				\hdashline

			    \textcolor{white}{w/o}w/o Sub-summary                        & 53.8 & 28.3 & 44.1 & 53.5 \\
			    \textcolor{white}{w/o}w/o Coherence                      & 53.9 & 28.6 & 44.2 & 53.5 \\
				\bottomrule
		\end{tabular}}
	    \vspace{-2mm}
		\caption{Results on SAMSum test split. $\ast$ indicates that the results are significantly different from ours ($p<0.05$).}
		\label{table:samsum-result}
		\vspace{-3mm}
	\end{table}

	\paragraph{Baselines} Lead3 is a commonly adopted method in the news summarization task, which simply takes the first three leading sentences of text as its summary.
	PTGen \citep{see2017get} extends sequence-to-sequence model with copy and coverage mechanisms.
	FastAbs-RL \citep{chen2018fast} first selects pivot sentences and then generates abstract summary with reinforcement learning.
	DynamicConv + GPT-2/News \citep{wu2018pay} proposes a lightweight dynamic convolutions to replace the self-attention modules in the Transformer layers.
	UniLM \cite{li2019unilm} is a unified language model which can be used for both natural language understanding and generation tasks.
	BART \citep{lewis2020bart} is a pre-trained encoder-decoder Transformer model, with two versions $\text{BART}_\textsc{base}$ and $\text{BART}_\textsc{large}$.
	For simplicity, we use BART to denote $\text{BART}_\textsc{large}$.
	Multiview BART \citep{chen2020multi} incorporates mutli-view features to summarize dialogues, including global, discrete, topic and stage information of dialogues.
	$\text{BART}_\textsc{ori}$ finetunes the $\text{BART}_\text{LARGE}$ with its original pre-training tasks (i.e., sentence shuffling and text infilling) \cite{lewis2020bart}, acted as auxiliary tasks like this work.

    \begin{table}[t]
		\centering
		\scalebox{0.85}{
		\begin{tabular}{lccccc}
			\toprule
			{Model} & {R-1} & {R-2} & {R-L}&\textsc{BertS}\\
			\midrule
			$\ast$Lead3                              & 15.0 & 5.1  & 13.3 & - \\
			$\ast$PTGen                    & 28.8 & 12.2 & 24.2 & - \\
			$\ast$UniLM                                & 32.7 & 17.3 & 29.8 & -  \\
			\midrule

			$\ast$$\text{BART}$           & 34.7 & 17.7 & 30.9 & 30.7 \\ 
			
			  $\ast$$\text{BART}_\textsc{ori}$    & 35.0 & 17.9 & 31.1 & 31.2 \\ 
			\textsc{ConDigSum}                                & \textbf{36.0} & \textbf{18.9 }& \textbf{32.2 }& \textbf{32.4} \\
			\hdashline
			\textcolor{white}{w/o}w/o Sub-summary                   & 35.5 & 18.7 & 31.9 & 32.0 \\
			\textcolor{white}{w/o}w/o Coherence                     & 35.5 & 18.6 & 31.7 & 31.9 \\
			\bottomrule
		\end{tabular}}
	    \vspace{-2mm}
		\caption{Results on MediaSum test split. $\ast$ indicates that the results are significantly different from ours ($p<0.05$).}
		\label{table:mediasum-result}
		\vspace{-3mm}
	\end{table}

	\paragraph{Results on SAMSum.} The results on SAMSum dataset are listed in Table~\ref{table:samsum-result}. 
	Results of Lead3, PTGen, DynamicConv + GPT-2/News, and FastAbs-RL are taken from \citealp{gliwa19}.
	Others are based on our implementations (see the appendix).
	As we can see that, according to ROUGE script our model \textsc{ConDigSum} significantly outperforms previous state-of-the-art models in the first block ($p<0.05$), indicated by $\ast$, with regard to both ROUGE and BERT scores, which demonstrates the effectiveness of the proposed contrastive objectives.
	Comparing $\text{BART}_\text{LARGE}$ against $\text{BART}_\textsc{ori}$, it is interesting to observe that treating the original pre-training objectives as auxiliary tasks during fine-tuning also leads to performance gains.
	However, our proposed contrastive objectives are more effective.

	We also conducted an ablation study on the SAMSum dataset.
	The ROUGE-2 score drops 0.7 points after removing the coherence detection objective, while the performance drops 1 point by ignoring the sub-summary generation objective.
	Such a phenomenon indicates both proposed contrastive objectives help generate better summaries, while the sub-summary generation objective contributes more to the primary task, compared to the coherence detection objective.
	One reason is that the sub-summary generation objective and the primary summary task are both sequence-to-sequence learning problems, yet the coherence detection objective only affects the encoder part. 
	
 	\paragraph{Results on MediaSum.} Table~\ref{table:mediasum-result} shows the results on the MediaSum dataset. 
	Results of PTGen and UniLM are reported by \citealp{zhu2021mediasum}.
	Similar to SAMSum, \textsc{ConDigSum} also outperforms all the baseline models. The ablation study on the MediaSum dataset shows both auxiliary tasks contribute to the primary task and the results of them are similar. 

	\paragraph{Impact of Different Multi-Task Combination Strategies.} 
	Table~\ref{table:altersum-large} listed the performance on SAMSum dataset adopting either the alternating parameter updating or the summation objective strategy.
	Compared to the $\text{BART}_\textsc{large}$ baseline in Table \ref{table:samsum-result}, both strategies result in performance gains, while the alternating parameter updating strategy is more helpful.
	Hence, this work adopts the alternating parameter updating strategy. 
	
	\begin{table}[t]
		\centering
		\scalebox{0.88}{
		\begin{tabular}{lcccc}
			\toprule
			{Mechanism} & {R-1} & {R-2} & {R-L} & {\textsc{BertS} }\\
			\midrule
			Alternating updating                  & 54.3 & 29.3 & 45.2 & 54.0 \\
			Summation objective                       & 53.3 & 28.2 & 44.1 & 53.1 \\
			\bottomrule
		\end{tabular}}
	    \vspace{-2mm}
		\caption{Results of multi-task combination strategies.}
		\label{table:altersum-large}
		\vspace{-3mm}
	\end{table}

	\begin{table}[t]
		\centering
		\scalebox{0.85}{
			\begin{tabular}{lccccc}
				\toprule
				Systems          & 1st & 2nd & 3rd & 4th  & MR \\
				\midrule
				BART         & 0.14     & 0.12    & 0.31    & 0.43       & 3.03     \\
				Multiview BART & 0.19    & 0.27    & 0.25    & 0.29       & 2.64    \\
				\textsc{ConDigSum}     & 0.26    & 0.32    & 0.23    & 0.19       & 2.35     \\
				\hdashline
				Gold             & 0.41    & 0.29    & 0.21    & 0.09        & 1.98     \\
				\bottomrule
		\end{tabular}}
		\caption{Human evaluation on SAMSum: proportions of rankings. MR: mean rank (the lower the better).} 
		\label{tab:human}
	\end{table}
	\paragraph{Human Evaluation.}
	Since the automatic evaluation mainly focuses on the semantic matching between the generated output and the ground truth, while the generated summaries may be disfluent or ungrammatical, we thus also elicit feedback from human efforts.
	We compared our proposed model with the human references, as well as two baselines, BART \cite{lewis2020bart} and Multiview BART \cite{chen2020multi}\footnote{Outputs are publicly available at \href{https://github.com/GT-SALT/Multi-View-Seq2Seq}{https://github.com/GT-SALT/Multi-View-Seq2Seq}}.
	100 dialogues are randomly selected from the test split of SAMSum dataset.
	10 participants are presented with a dialogue and its paired candidate summaries, including human references, generated outputs by three models.
	For each selected dialogue, they are asked to rank the candidate output from the best to worst with regard to \textit{fluency} (is the summary fluent/grammatically correct?), \textit{informativeness} (does the summary contains the most informative pieces of the dialogue?), and \textit{succinctness} (does the summary express in an abstractive way?). 
	Table \ref{tab:human} listed the proportions of different system rankings and mean rank (lower is better).
	The output of our \textsc{ConDigSum} is ranked as the most appropriate summary for 26\% of all cases.
	Overall, we obtain lower mean rank than the other two systems but still lags behind the Gold one.

	\subsection{Case Study and analysis}
	\begin{figure}[t]
        \centering
        \vspace{+3mm}
        \includegraphics[width=0.42\textwidth]{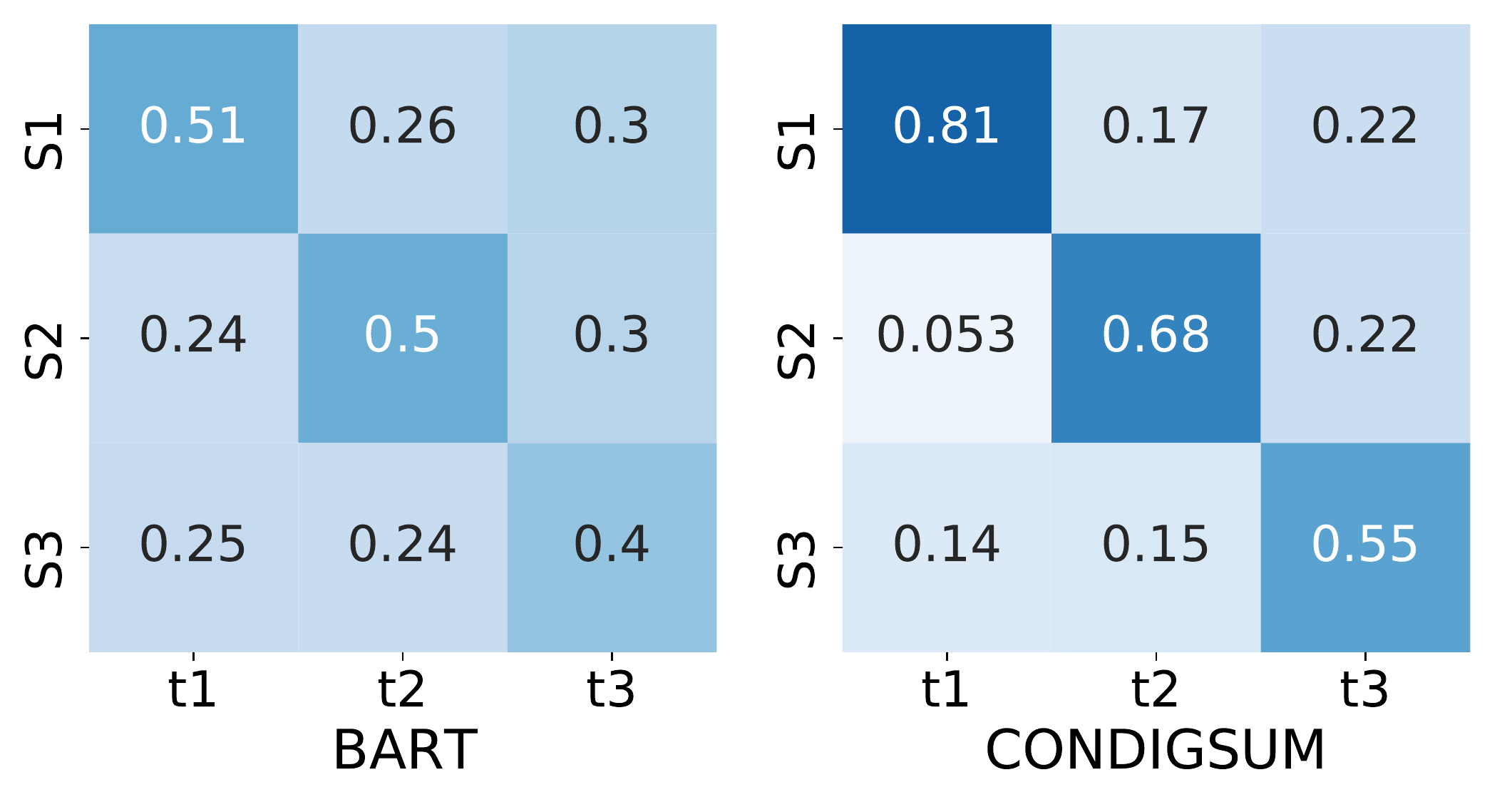}
        \caption{Visualization of how much a sub-summary is related to different snippets (the sum of every column is equal to 1). The result of \textsc{ConDigSum} is more concentrated on diagonal. }
        \vspace{-2mm}
        \label{Fig:case-visual}
        \vspace{-3mm}
    \end{figure}

	\paragraph{How do coherence and sub-summary objectives work?}
	Firstly, we compared the coherence scores predicted by our \textsc{ConDigSum} model of intra-topic snippets and inter-topic snippets. 
	Taking the dialogue in Figure \ref{Fig:intro-example} from the test split of SAMSum dataset as an example, coherence scores of intra-topic snippets $S_1,S_2$ and $S_3$ are 1.37, 2.17 and 3.12, respectively, while the scores of inter-topic snippets $S_4$ and $S_5$ are much lower (-0.15 and -5.64, individually).\footnote{Refer to the appendix for the illustration.} 
	This indicates that the coherence detection objective does help the model capture the topical information of the dialogue. 
	On the other hand, we tried to find out how the sub-summary generation objective affects the generation of summaries. For the same dialogue, we calculated the sequence-to-sequence loss of snippet-summary pairs \{($ S_i $, $ t_j $), $ i, j \in \{1, 2, 3\} $ \} by feeding each snippet-summary pair into the trained model (the snippet for encoder and the summary for decoder).
	The log-likelihood loss was then transformed to represent the correlation score between a snippet and a sub-summary (a lower loss means a higher correlation). 
	Figure \ref{Fig:case-visual} visualizes how much one sub-summary is related to different snippets (i.e., every column). The results of our \textsc{ConDigSum} model were more concentrated on the diagonal than those of BART, which proves that our sub-summary generation objective indeed forces the model to pay more attention to the most salient fact and generate more relevant summaries. 
	
	\begin{figure}[t]
        \centering
        \vspace{+3mm}
        \includegraphics[width=0.48\textwidth]{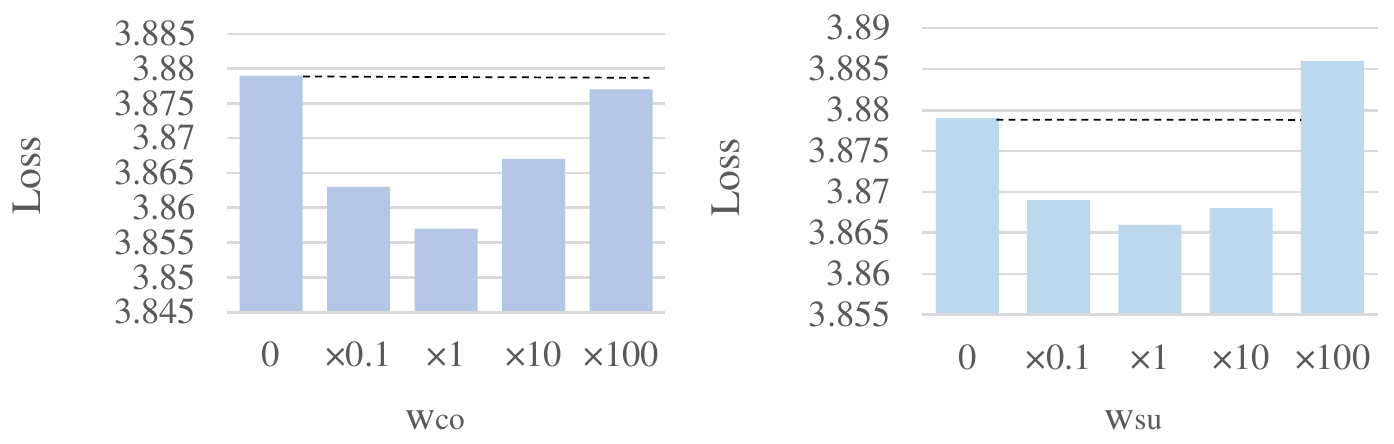}
        \caption{Impact of different values of task-coefficients of coherence detection (left) and sub-summary generation (right) objectives on the validation loss of the primary dialogue summarization task. }
        \vspace{-2mm}
        \label{Fig:weightimpact}
        \vspace{-3mm}
    \end{figure}
	\subsection{How does task-wise coefficients affect primary task?} In order to make it easier to observe 
	how the task-wise coefficients affect the primary task, we only consider one contrastive objective at each time, by removing either the sub-summary generation objective or the coherence detection objective as well as scaling down and up the optimal values of the task-coefficients, $w_{co}$ and $w_{su}$, based on the optimal values (denoted as $\times 1$).
	The values of the primary summrization loss on SAMSum dataset with different task-coefficients are depicted in Figure \ref{Fig:weightimpact}.
    We can observe that the primary loss increases with either larger or smaller task-coefficients.
    Assigning larger weights to the auxiliary tasks will encourage the model to prefer auxiliary tasks and ignore the primary task, where the primary task converges to the sub-optimal point. 
    However, auxiliary tasks assigned by too small weight numbers will fail to assist the model to capture the dialogue topic information. 
	
	\paragraph{Is the coherence detection objective actually topical-related? }
    To quickly investigate the relationship between coherence detection objective and discourse structures, we constructed a set of 70 contrastive examples. Each example is constructed as follows: For a snippet $ s_1 $, consisting of utterances from the same topic in a dialogue, we randomly select an utterance u in $ s_1 $ and replace it with another utterance v from other topics, where the dialogue act types for u and v are the same. Therefore, we get a new snippet $ s_2 $. The encoder of our model is used to get the coherence scores for $ s_1 $ and $ s_2 $, respectively. We found that the average coherence scores(-0.73) for the original snippets($ s_1 $) are higher than the scores for their counterparts($ s_2 $) with replacements(-1.02). Though the two examples have the same dialogue act types, the coherence scores are different. From this, we think the coherence detection objective does capture topic-related information. Moreover, from our understanding, a dialogue’s topic and its discourse structure can be interlaced. The coherence score distribution of dialogue can reflect the topic change and also correlate to the discourse flow, while our work mainly focuses on the first point. 
	
\begin{figure}[t]
        \centering
        \includegraphics[width=0.42\textwidth]{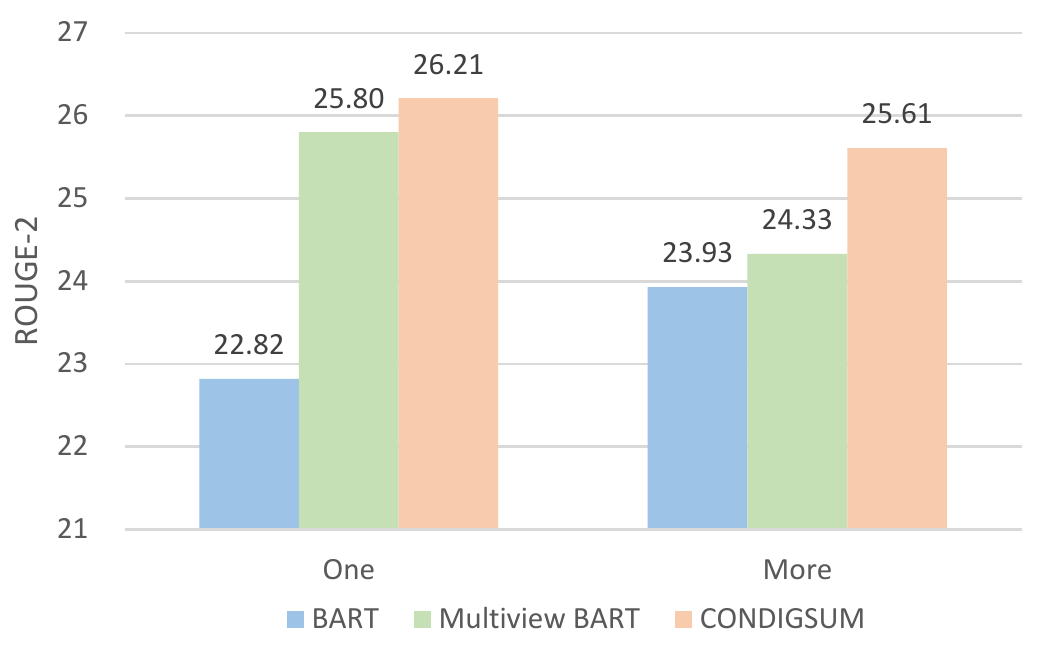}
        \caption{ROUGE-2 score of generated summaries for dialogues containing one or more sub-summaries. }
        \vspace{-2mm}
        \label{Fig:case-rouge}
        \vspace{-3mm}
    \end{figure}

	\paragraph{Relation between the quality of summary and complexity of dialogues.} We further investigated the relation between the quality of generated summaries with regard to the number of sub-summaries residing in a dialogue summary. 
	The test split of SAMSum dataset was divided into two sets: a) \textbf{One}: the dialogue summaries that only contain one sub-summary; b) \textbf{More}: the dialogue summaries consisting of more than one sub-summaries.
	For each set, we calculated the averaged ROUGE-2 score over all elements.
	We include \textsc{ConDigSum}, BART \cite{lewis2020bart} and Multiview BART \cite{chen2020multi} for comparison, as listed in Figure~\ref{Fig:case-rouge}.
    Our model performs better than two baselines under both circumstances. In addition, under the \textbf{One} situation, \textsc{ConDigSum} outperforms Multiview BART by 0.41 ROUGE-2 point, yet the difference is expanded to 1.28 points under \textbf{More}. 
    This increment indicates that our model significantly improves the quality of generated summaries when the dialogue summary comprises of more than one sub-summaries.

	\section{Related Work}
	
	\paragraph{Document Summarization.} 
	Automatic document summarization aims to condense a well-structured document into its shorter form where the important information preserved. 
	This task can be categorized into extractive  and abstractive document summarization. The extractive summarizer learns to find the informative sentences from the input document as its summary, which can be viewed as a sentence problem \cite{kupiec1995trainable,conroy2001text}.
	The features can be learned from LSTMs, CNNs or Transformers	 \cite{cheng2016neural,Nallapati2017SummaRuNNer,zhang2018neural,zhang2019hibert,liu2019text}.
	The abstractive summarization task learns to generate summaries by rewriting the input document, which is a typical sequence-to-sequence learning problem.
	Sequence-to-sequence attentive LSTMs \cite{hochreiter1997long,bahdanau2015neural} and its extensions with copy mechanism \cite{gu2016incorporating}, coverage mechanism \cite{see2017get} and reinforcement learning \cite{paulus2018deep} have shown effectiveness on summarizing the document. 
	Recent studies have investigated the pretrained transformer models, like BERTAbs \cite{liu2019text}, BART \cite{lewis2020bart}, PEGASUS \cite{zhang2020pegasus} and STEP \cite{zou2020pre}.
	
	The extractive and abstractive methods can be combined with reinforcement learning \cite{chen2018fast}, attention mechanisms \cite{gehrmann2018bottom,hsu2018unified} or in a pipeline manner \cite{pilault2020extractive}, while this work focuses on summarizing dialogue utterances from a sequence-to-sequence learning perspective.
	
	\paragraph{Dialogue Summarization.} 
	The dialogue summarization task aims to summarize the dialogue content consisting of utterances from multiple speakers.
	\citet{shang2018unsupervised} proposed a simple multi-sentence compression technique to summarize meetings in an unsupervised fashion.
	\citet{zhao2019abstractive,zhu2020end} designed hierarchical model structures to capture features of conversational utterances from different turns.
	
	The conversational analysis can also be unitized to generate the summaries for dialogue content.
	\citet{Liu2019TopicAwarePN,li2019keep} introduced the topical information to the summarization process, while \citet{liu-chen-2019-reading} took use of the key utterances and \citet{goo2018abstractive} leveraged the dialogue acts.
	\citet{chen2020multi} explicitly modeled conversational structures from four different views and then design a multi-view decoder to incorporate features from such four views to generate dialogue summaries. 
	However, the additional information of conversational topics, key utterances, dialogue acts, and conversational structures requires human annotations, which is quite expensive or requires extra segment algorithms.
	Without requiring extra human effort or algorithms, this work proposes to introduce two contrastive learning objectives as auxiliary tasks during training.
	
	\paragraph{Contrastive Learning.} 
	The application of contrastive learning for various tasks has been investigated recently, mainly in computer vision domain.
	The contrastive predictive coding \cite{oord2018representation} has been studied for data-efficient image recognition \cite{henaff2020data}.
	Without using specialized architectures or a memory bank, learning visual representations in a contrastive manner outperforms various baselines with self-supervised, semi-supervised and transfer learning \cite{chen2020simple}.
	\citet{khosla2020supervised} proposed a fully-supervised contrastive loss which achieved new state-of-the-art results on the image classification task, surpassing the cross-entropy loss.
	This work also demonstrates that, compared to the traditional cross-entropy loss,  the proposed supervised contrastive loss performs more stably to different hyperparameter settings, like data augmentations and optimizers. 
	Moreover, \citet{klein2020contrastive} introduced contrastive margin as regularizer for commonsense reasoning where a pairwise contrastive auxiliary prediction task is constructed. 
	\citet{fang2020cert} proposed to pre-train language models with contrastive self-supervised learning at the sentence level, which learns to predict whether two sentences originate the same one. 
	\citet{gunel2020supervised} proposed a supervised contrastive learning objective which allows to work with cross-entropy and lead to significant performance gains.
	The contrastive learning is also introduced to learn the sentence embeddings \cite{gao2021simcse}.
	The above applications of contrastive learning are for computer vision or natural language understanding domains, while, in this work, we introduce the contrastive learning to the abstractive dialogue summarization task, which is a typical generation task.
	
	\section{Conclusion}
	Recent research progresses have present the effectiveness of dialogue studies (e.g., topical information and dialogue acts) on summarizing dialogues, while additional expert annotations or extra algorithms are required to obtain the knowledge.
	This work proposes a simple yet effective method, \textsc{ConDigSum}, that implicitly captures the topical knowledge residing in dialogue content by modeling the text coherence, yet no additional human annotations or segment algorithms are needed.
	We design two contrastive objectives as auxiliary task, i.e., coherence detection and sub-summary generation objectives,working together with the primary summarization task during training.
	An alternating parameter update strategy is employed to cooperate the primary and auxiliary tasks.
	Experiments on two benchmark datasets demonstrate the efficacy of the proposed model.
	Future directions include learning structured representations of information flow residing in dialogues and leveraging knowledge graphs to generate better dialogue summaries.

	\section{Ethical Considerations}
	Our simple yet effective abstractive dialogue summarization system could be used where there exists dialogue systems (two or multi-party dialogues). For example, it could be used for grasping the key points quickly or recapping on the salient information of online office meeting. In addition, the system can also be used for customer service, requiring employees to summarize the conversation records of customers' inquiries, complaints and suggestions. 
	
	The daily dialogue and media interview datasets used in this work are publicly available, and only for research purpose. There may exist biased views in them, and the content of them should be viewed with discretion. 
	
    \section*{Acknowledgements}
    Yanyan Zou and Xiaojie Wang are the corresponding authors. We would like to thank anonymous reviewers for their suggestions and comments. The work was supported by the National Natural Science Foundation of China (NSFC62076032) and the Cooperation Project with Beijing SanKuai Technology Co., Ltd. We would also like to thank all annotators who have contributed to the case study, especially Rizhongtian Lu. 
	
	
	\bibliography{emnlp2021}
	\bibliographystyle{acl_natbib}
	
	\clearpage
\appendix
\section{Experiment}
		
	\begin{table*}[t]
		\centering
		\begin{tabular}{lccccc}
			\toprule
			{Datasets} & {DialogToken} & {UtterToken} & {SummaryToken} & {DialogUtter} & {Speaker} \\
			\midrule
			SAMSum     & 93.8   & 9.5  & 20.3 & 9.9  & 2.2 \\
			MediaSum   & 1,553.7 & 51.7 & 14.4 & 30.0 & 9.2 \\
			\bottomrule
		\end{tabular}
		\caption{Data statistics of dialogue summarization datasets. \textit{DialogToken}, \textit{UtterToken} and \textit{SummaryToken} stand for the average number of tokens in dialogues, utterances and summaries, respectively. \textit{DialogUtter} is the average number of utterances in dialogues. The last column lists the average number of speakers in dialogues. }
		\label{table:datasets}
        \vspace{-3mm}
    \end{table*}
	
	\subsection{Dataset}
	
	We also show detailed statistics about such two datasets, SAMSum \cite{gliwa19} and MediaSum \cite{zhu2021mediasum}, with regard to average tokens, utterances and speakers, as showed in Table~\ref{table:datasets}.
	It is straightforward that the dialogue in MediaSum is much longer than the one in SAMSum, yet the corresponding summary is much shorter.
	
	\subsection{Implementation Details}
	The sequence-to-sequence Transformer model is adopted as our backbone architecture, implemented using Fairseq toolkit\footnote{We empirically observed that different frameworks (e.g. Fairseq and Huggingface Transformer) may obtain different results under the same hyperparameter settings.} \citep{ott2019fairseq}. 
	To be specific, our model is initialized with a pre-trained sequence-to-sequence, i.e., BART \cite{lewis2020bart}.
	Thus they share the same architectures, 6-layer encoder-decoder Transformer for $\text{BART}_\text{BASE}$ and 12-layer Transformer for $\text{BART}_\text{LARGE}$.
	Each layer in $\text{BART}_\text{BASE}$ has 16 attention heads, and the hidden size and feed-forward filter size is 1024 and 4096, respectively, resulting in 140M trainable parameters. 	
	Each layer in $\text{BART}_\text{LARGE}$ has 16 attention heads, and the hidden size and feed-forward filter size is 1024 and 4096, respectively, resulting in 400M trainable parameters. 
	The dropout rates for all layers are set to 0.1.
	The optimizer is Adam \cite{kingmaadam} with warmup.
 
	For SAMSum dataset, the learning rate $\alpha$ is 4e-5, and the maximum number of tokens in each batch is 800. 
	The model is trained for 3 epochs.
	Each epoch takes around 0.7 hours on single Tesla P40 GPU.
	The window size $k$ of the coherence detection objective is tuned over 5 to 15, with a stride of 2. 
	The optimal value is 14.
	The lower bound of sliding window size for the sub-summary generation objective is selected from [1, 5], with the difference between lower and upper bounds set to 20. 
	The optimal values for $w_{co}$ and $w_{su}$ are 0.005 and 0.0001 individually. The number of contrastive pairs for each sample, i.e., $N_{co}$ and $N_{su}$, is equal to 2.
	
	For MediaSum dataset, the learning rate $\alpha$ is 2e-5, and the maximum number of tokens in one batch is 1100. 
	The model is trained for 4 epochs, each of which takes around 15 hours on four Tesla V100 GPUs.
	Similar to SAMSum, for the coherence detection objective, the window size $k$ is 10, and the task-wise coefficient $w_{co}$ is 0.00005.
	The sliding window size interval of the sub-summary generation objective is $[1, 5]$, with the task-wise coefficient $w_{su}$ of 0.00005. For simplicity, the number of contrastive pairs for each sample, i.e., $N_{co}$ and $N_{su}$, is equal to 1.
    Following \citet{zhu2021mediasum}, we add interlocutors information before concatenating utterances, and then truncate the dialogues to keep only first 1024 tokens as input. 
  	All experiments were conducted on either Tesla P40 GPUs (24GB) or Tesla V100 GPUs (16GB).
  	
  	\subsection{Construction of Sub-summary}
  	All sub-summaries are constructed from ground-truth summaries following a pre-processing procedure. 
  	We only consider dialogues whose ground-truth summary consists of at least two sentences and filtered the sentences in ground-truth summaries that have no good match with any snippets in original dialogues in terms of ROUGE score.
  	We also tried to take BertScore as the selection metric of snippets, but ROUGE was finally adopted because there is barely any difference between them and the cost of BertScore was much larger.

    \subsection{Results}
    The output of MultiviewBART \cite{chen2020multi} is publicly available at \href{https://github.com/GT-SALT/Multi-View-Seq2Seq}{https://github.com/GT-SALT/Multi-View-Seq2Seq}.
    Since the ROUGE scores may vary due to different toolkits, to make fair comparisons with our model, we recalculated the ROUGE scores on the output of MultiviewBART using the files2rouge\footnote{\href{https://github.com/pltrdy/files2rouge}{https://github.com/pltrdy/files2rouge}}, same as ours. 
    
	\begin{figure*}[t]
        \centering
        \includegraphics[width=0.85\textwidth]{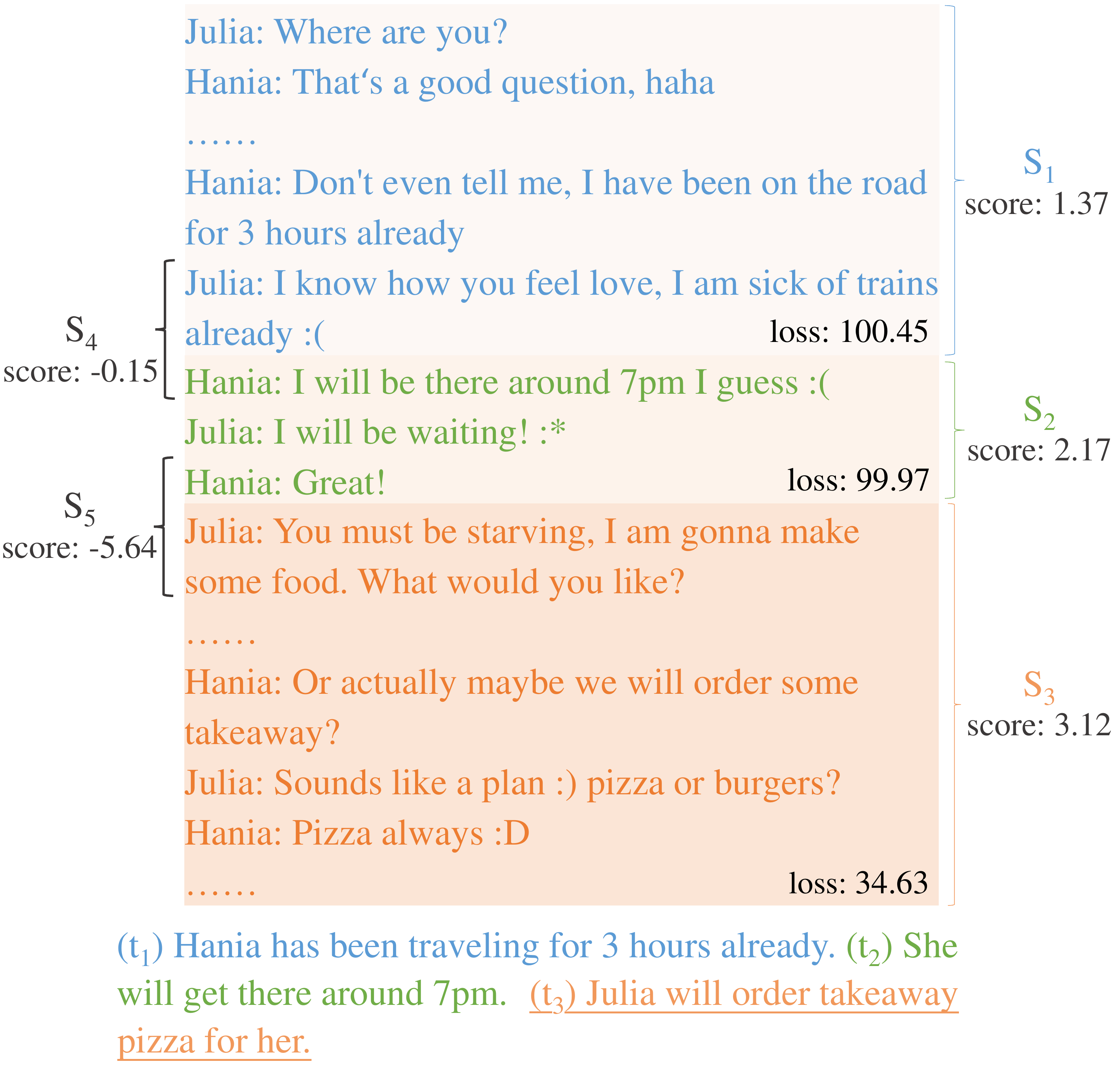}
        \caption{Coherence scores of snippets and visualization of how the sub-summary $t_3$ is related to different snippets $S_i$ ($i \in \{1, 2, 3\}$). Darker background means a smaller loss and higher correlation between one snippet and the sub-summary $t_3$. }
        \vspace{-2mm}
        \label{Fig:case-appendix}
        \vspace{-1mm}
    \end{figure*}
    
    \subsection{Performance on the Validation Set}
    The performance on the validation split of SAMSum and MediaSum is listed in Table \ref{table:samsum-valid} and \ref{table:mediasum-valid}, respectively.
    
     \begin{table}[t]
		\centering
		\scalebox{1}{
			\begin{tabular}{lcccc}
				\toprule
				{Model} & {R-1} & {R-2} & {R-L} \\
				\midrule
				$\text{BART}_\textsc{base}$     & 48.7 & 25.2 & 39.1 \\
				$\text{BART}$                   & 54.0 & 28.8 & 44.0 \\ 
				$\text{BART}_\textsc{ori}$      & 53.7 & 28.2 & 43.5 \\  
                $\textsc{ConDigSum}_\text{BASE}$& 50.7 & 26.9 & 41.6 \\
				\textsc{ConDigSum}              & 55.3 & 30.5 & 45.5 \\
                \hdashline
			    \textcolor{white}{w/o}w/o Sub-summary       & 54.9 & 29.5 & 44.6 \\
			    \textcolor{white}{w/o}w/o Coherence         & 54.8 & 29.6 & 44.9 \\
				\bottomrule
		\end{tabular}}
		\vspace{-2mm}
		\caption{Results on SAMSum validation split.}
		\label{table:samsum-valid}
	    \vspace{-3mm}
	\end{table}
    	
    \begin{table}[t]
		\centering
		\scalebox{1}{
		\begin{tabular}{lcccc}
			\toprule
			{Model} & {R-1} & {R-2} & {R-L}\\
			\midrule
			$\text{BART}$           & 34.9 & 17.8 & 31.0  \\ 
			$\text{BART}_\textsc{ori}$  & 35.0 & 17.8 & 31.0  \\ 
			\textsc{ConDigSum}                                & 35.6 & 18.7 & 31.9 \\
			\hdashline
			\textcolor{white}{w/o}w/o Sub-summary       & 35.4 & 18.5 & 31.8 \\
			\textcolor{white}{w/o}w/o Coherence         & 35.3 & 18.4 & 31.6 \\
			\bottomrule
		\end{tabular}}
		 \vspace{-2mm}
		\caption{Results on MediaSum validation split.}
		\label{table:mediasum-valid}
		 \vspace{-3mm}
	\end{table}

    
    \subsection{Sensitivity tests}
    To explore the effects of the hyper-parameters of our methods, we conducted sensitivity tests on validation split of SAMSum. Generally, there is an optimal value reaching at highest ROUGE scores, while too small or too large values hamper performance. $ \ast $ indicates the best setting according to the validation set. 
    
    \begin{table}[t]
		\centering
		\scalebox{1}{
		\begin{tabular}{lcccc}
			\toprule
			{$ k $} & {R-1} & {R-2} & {R-L} \\
			\midrule
			$ 2 $           & 54.9 & 29.8 & 45.0  \\ 
			$ 4 $  & 54.8 & 29.6 & 44.8  \\ 
			$ 6 $  & 54.7 & 29.3 & 45.0  \\ 
			$ 8 $  & 54.8 & 29.5 & 44.9  \\ 
			$ 10 $  & 54.7 & 29.5 & 44.8  \\ 
			$ 14 \ast $  & 55.3 & 30.5 & 45.5  \\ 
			$ 18 $  & 54.6 & 29.6 & 45.0 \\
			\bottomrule
		\end{tabular}}
		\caption{Sensitivity test of the coherence window $ k $. }
		\label{table:sensitivity-k-valid}
	\end{table}
	
	\begin{table}[t]
		\centering
		\scalebox{1}{
		\begin{tabular}{lcccc}
			\toprule
			{$ a $} & {R-1} & {R-2} & {R-L} \\
			\midrule
			$ 1 $           & 54.4 & 29.2 & 44.6  \\ 
			$ 3 $  & 54.7 & 29.4 & 44.7  \\ 
			$ 5 \ast $  & 55.3 & 30.5 & 45.5 \\ 
			$ 7 $  & 54.8 & 29.3 & 44.9  \\ 
			\bottomrule
		\end{tabular}}
		\caption{Sensitivity test of the sub-summary window's lower bound $ a $. }
		\label{table:sensitivity-a-valid}
	\end{table}
	
	\begin{table}[t]
		\centering
		\scalebox{1}{
		\begin{tabular}{lcccc}
			\toprule
			{$ N_{co} $} & {R-1} & {R-2} & {R-L} \\
			\midrule
			$ 1 $           & 54.8 & 29.7 & 45.0  \\ 
			$ 2 \ast $  & 55.3 & 30.5 & 45.5  \\ 
			$ 3 $       & 55.0 & 29.8 & 45.3 \\
			\bottomrule
		\end{tabular}}
		\caption{Sensitivity test of the number of contrastive pairs for each sample $ N_{co} $. }
		\label{table:sensitivity-co-sample-valid}
	\end{table}
	
	\begin{table}[t]
		\centering
		\scalebox{1}{
		\begin{tabular}{lcccc}
			\toprule
			{$ N_{su} $} & {R-1} & {R-2} & {R-L} \\
			\midrule
			$ 1 $           & 54.9 & 29.8 & 45.0  \\ 
			$ 2 \ast $  & 55.3 & 30.5 & 45.5  \\ 
			$ 3 $       & 54.7 & 29.5 & 45.0 \\
			\bottomrule
		\end{tabular}}
		\caption{Sensitivity test of the number of contrastive pairs for each sample $ N_{su} $. }
		\label{table:sensitivity-ma-sample-valid}
	\end{table}



    \subsection{Case Study}
    A complete example showing coherence scores of different snippets and the generation loss of one sub-summary with respect to different snippets is shown in Figure \ref{Fig:case-appendix}. 
\end{document}